\documentclass[10pt, a4paper, conference, compsocconf]{IEEEtran}
\ifCLASSINFOpdf
  \usepackage[pdftex]{graphicx}
  \graphicspath{{figures/}}
  \DeclareGraphicsExtensions{.pdf,.jpeg,.png,.eps}
\else
  \usepackage[dvips]{graphicx}
  \graphicspath{{figures/}}
  \DeclareGraphicsExtensions{.pdf,.jpeg,.png}
\fi
\usepackage[caption=false,font=footnotesize]{subfig}
\usepackage{stfloats}
\begin{document}
\title{Symbol detection in online handwritten graphics \\
using Faster R-CNN}


% \author{\IEEEauthorblockN{Authors Name/s per 1st Affiliation (Author)}
% \IEEEauthorblockA{line 1 (of Affiliation): dept. name of organization\\
% line 2: name of organization, acronyms acceptable\\
% line 3: City, Country\\
% line 4: Email: name@xyz.com}
% \and
% \IEEEauthorblockN{Authors Name/s per 2nd Affiliation (Author)}
% \IEEEauthorblockA{line 1 (of Affiliation): dept. name of organization\\
% line 2: name of organization, acronyms acceptable\\
% line 3: City, Country\\
% line 4: Email: name@xyz.com}
% }

\author{\IEEEauthorblockN{Frank D. Julca-Aguilar and Nina S. T. Hirata}
   \IEEEauthorblockA{Department of Computer Science, Institute of
     Mathematics and Statistics\\
     University of S\~ao Paulo (USP)\\
     S\~ao Paulo, Brazil}
}


%Telephone: (800) 555--1212, Fax: (888) 555--1212}


\maketitle

\begin{abstract}
Symbol detection techniques in online handwritten graphics 
(e.g. diagrams and mathematical expressions)
consist of methods specifically designed for 
a single graphic type. In this work, we evaluate the Faster R-CNN 
object detection algorithm as a general method  for detection 
of symbols in handwritten graphics. 
We evaluate different configurations of 
the Faster R-CNN method, and point out issues 
relative to the handwritten nature of the data. 
Considering the online recognition context, we 
evaluate efficiency and accuracy trade-offs of using  
Deep Neural Networks of different complexities 
as feature extractors.
We evaluate the method on publicly available
flowchart and mathematical expression (CROHME-2016)
datasets. Results show that Faster R-CNN can be effectively 
used on both datasets, enabling the possibility 
of developing general methods for symbol detection, and 
furthermore, general graphic understanding methods 
that could be built on top of the algorithm.
% We obtained encouraging results not only considering
% symbol detection, but for general graphics understanding methods 
% that can be built on top of the proposed models. We release 
% the code and models as open source.
% We obtained X mAP@0.5 on the first dataset and 
% Y mAP@0.5 on the second. Results are encouraging not only 
% for symbol detection, but for general graphics understanding methods 
% that can be built on top of the proposed models. We release 
% the code and models as open source.

% By converting online data to offline (images),
% such the symbol segmentation and classification task
% can be cast as an object detection problem in images.
% Using Convolutional methods then can be used to 
% tackle recognition of symbols in different graphics domains 
% with a same framework.
% In this work, we evaluate the Faster R-CNN method
% for the recognition of symbols in mathematical expressions 
% and flowcharts.
% and evaluate the performance of the method over 
% different Deep Convolutional Neural Networks. 
% We evaluate the method in the recognition of handwritten
% mathematical expression and flowchart datasets.
% In flowcharts, we obtain
% We release the models as open source, which can 
% easily used to build complete graphic recognition methods 
% on top of them.   

\end{abstract}


% \fi
\IEEEpeerreviewmaketitle

\section{Introduction}
\begin{figure}[!htb]
\centering
% \subfloat[]{\includegraphics[width=0.7\linewidth]{math_raw}} \\
% \subfloat[]{\includegraphics[width=0.7\linewidth]{math_img}}
\includegraphics[width=0.9\linewidth]{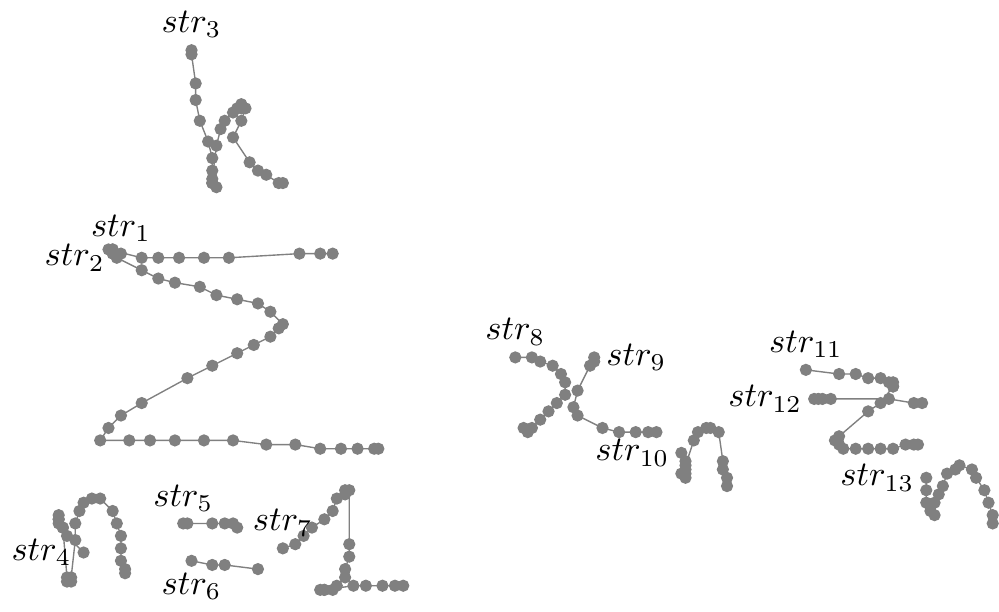}
\caption{Online handwritten mathematical expression 
composed of strokes $\{str_1, \ldots, str_{13}\}$.
Coordinates are depicted as gray circles.}
\label{fig:input}
\end{figure}

An online handwritten graphic is composed of a set 
of strokes, where each stroke consists of a 
set of bidimensional coordinates. The coordinates 
can be captured, for example, using a device with 
touch screen and an electronic pen. A symbol consists of 
a subset of strokes. In these data,
in contrast to text, symbols might be placed over vertical 
or diagonal positions relative to each other. 
Figure~\ref{fig:input} shows an online handwritten 
mathematical expression example.

% \begin{figure}[h]
% 	\centering
% \resizebox{0.6\linewidth}{!}{\input{images/raw_expression_sample.tikz}}

% \label{fig:raw_math_expression}
% %\caption{Handwritten version of  
% % $\sum \limits_{n=1}^{k} x_n z_n$, extracted from CROHME-2014 dataset\footnote{Competition on Recognition of Online Handwritten Mathematical Expressions}.}
% \end{figure}

Typical symbol detection techniques for online 
handwritten graphics include stroke grouping 
and classification processes. The first process selects
groups of strokes that are likely to form symbols, 
and the second applies machine learning methods 
to classify the stroke groups as symbols, 
with their corresponding symbol classes, or as
false positives. Due to the variance of the placement of 
symbols, virtually any group of strokes might form a symbol.
To avoid the computational complexity of evaluating all 
possible stroke groups, constraints based on handcrafted rules 
(e.g. only selecting stroke groups of up to four strokes)
are applied. Such constraints do not 
only limit the accuracy of the methods, but also make 
difficult the application of a same method to the
recognition of different graphic types.

Taking advantage of deep convolutional neural 
networks (DCNNs), recent algorithms for object detection 
have obtained 
outstanding accuracy. Among the different 
methods, Faster R-CNN has shown to obtain 
state-of-the-art accuracy and efficiency~\cite{Huang:2017}. 
Also, Faster R-CNN models are general enough to be applied 
to a large variety of problems as they can 
be trained end-to-end using input-output examples.

By converting raw online graphics data to offline (i.e. images),
object detection methods based on DCNNs could also be 
applied to symbol detection.
In this work, we evaluate the Faster R-CNN algorithm 
to detect symbols in handwritten graphics. 
We make a parallel between traditional symbol 
detection methods in online data and our methods 
(Section \ref{sec:related}). We then describe our pipeline to 
transform online data to offline, and give an overview of 
the Faster R-CNN algorithm (Section~\ref{sec:methods}). 
Through experimentation in the detection of 
symbols in mathematical expressions and flowcharts 
(Section \ref{sec:experimentation}), we show that the 
Faster R-CNN algorithm provides high accuracy on both 
problems. Results are encouraging not only for the 
development of general methods for symbol detection, but 
also for the development of methods for structure recognition 
(Section~\ref{sec:conclusions}). The code implemented in 
this work is available as open source.

% We analyze efficiency and effectiveness aspects of the algorithm, 
% pointing out issues relative to the handwritten recognition data.
% We provide open source code and models for further work.

\section{Related work}
\label{sec:related}
\begin{figure}[!htb]
\centering
\includegraphics[width=0.9\linewidth]{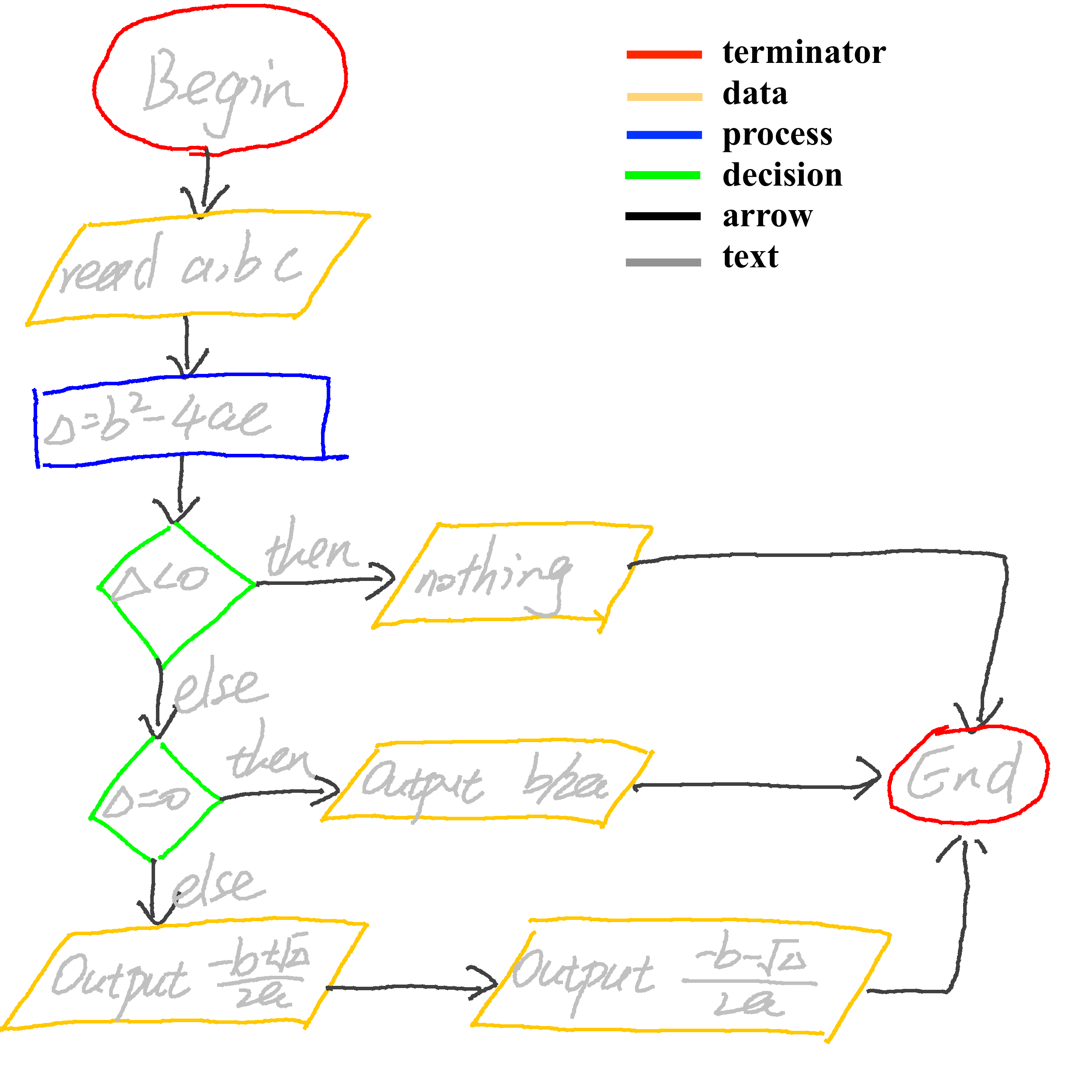}
\caption{Handwritten flowchart example.}
\label{fig:flowchart}
\end{figure}

We can find a variety of techniques for detecting 
symbols in online handwritten graphics. 
Most techniques introduce constraints based on 
some characteristics of the graphic type. 
For instance, in mathematical expression recognition, 
stroke grouping is often done considering only 
groups of strokes that have up to four 
or five strokes~\cite{Alvaro:2014,Frank:2014a,Awal:2012}.
Other common constraints include the assumption 
that symbols are formed only by strokes consecutive 
in input time order~\cite{Huang:2007,Lehmberg:1996}, or strokes 
that intersect each other~\cite{Tapia:2004}.
In the recognition of other graphic types, as diagrams, different techniques are 
designed to detect specific symbol 
classes~\cite{Bresler:2013,Bresler:2014,Carton:2013}. For instance, 
\textit{Bresler et. al.}~\cite{Bresler:2014} 
separate the detection of symbols that do not have 
an specific shape, as \textit{text} and \textit{arrows}, 
from symbols that have well defined shapes, as \textit{decision} 
and \textit{data}~\cite{Bresler:2014} (flowchart symbol examples 
are shown in Figure~\ref{fig:flowchart}).

Recent works on object detection are mainly based 
on DCNNs~\cite{Huang:2017}. At a high level, 
several of those 
techniques (e.g.~\cite{Liu:2016,Ren:2015,Girshick:2014,Girshick:2015})
consist of three processes: 
feature maps generation using a DCNN, selection of 
object bounding box candidates, and the classification 
of the bounding box candidates using the feature maps
(cropped according to the box dimensions). 
One of the algorithms that implements the above 
methods is Faster R-CNN. The algorithm has obtained 
state-of-the-art accuracy~\cite{Huang:2017} and has the 
advantage of doing the three processes through a single 
forward pass of a network.

While methods for symbol detection in online data 
are usually evaluated at stroke level~\cite{chrome:2016}, 
detection methods are 
evaluated at bounding box level (e.g. using mean average 
precision~\cite{Everingham:2010}). The evaluating metrics 
are then not directly comparable.

Although it could be possible to develop algorithms 
to recover stroke level information from the offline data, 
to the end of graphics understanding, such process might 
not be necessary. For instance, to recognize flowcharts
structure, once symbol candidates have been identified, 
relations between symbols can be determined using 
features from the corresponding bounding box 
regions over an image.

% Detection of symbols in an online version has a 
% different output. Although strokes could be recovered 
% with an additional process, to the end of understanding the 
% graphics, it might not be required to do that.

\section{Methods}
\label{sec:methods}
Faster R-CNN is a supervised learning algorithm.
The algorithm receives as input an image, and generates 
as output a list of object bounding box coordinates and the corresponding
object class per box. Training the algorithm then requires, 
in addition to the input images, a list of bounding boxes 
per image. This section gives details about 
the methods used to generate training data to 
evaluate the algorithm in the context of graphics recognition, 
and gives a brief description of the algorithm.
% This section describes the methods used to generate 
% the training data for Faster R-CNN algorithm and 
% gives a brief overview of the algorithm.

\subsection{Training data generation}
\begin{figure}[!t]
\centering
\subfloat[]{\fbox{\includegraphics[width=0.31\linewidth]{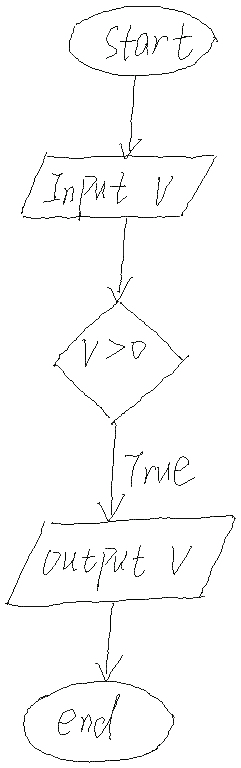}}}
\hfill
\subfloat[]{\fbox{\includegraphics[width=0.4\linewidth]{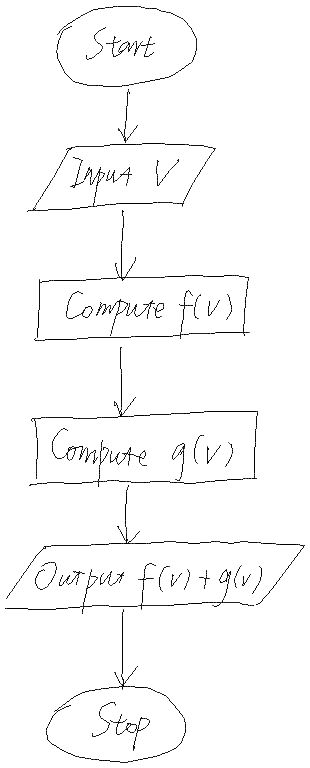}}} \\
% \subfloat[]{\includegraphics[width=0.8\linewidth]{inv-image-42}}
% \hfil
\subfloat[]{\fbox{\includegraphics[width=\linewidth]{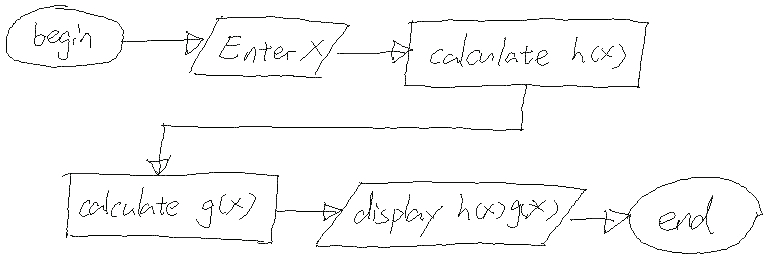}}}
\caption{Generated images with different 
aspect ratios.}
\label{fig:offline}
\end{figure}

% The training data consists of images of the handwritten 
% graphics and a list of object bounding boxes 
% (with their corresponding symbol class) per image.
% Figure X shows the pass of an two online math expressions 
% to their corresponding offline versions.
% Online handwritten data is composed of a sequence of strokes,
% each stroke being a sequence of two dimensional coordinates.
% The coordinates are captured, for example, through 
% a touch screen device and an electronic pen.
Depending on the input device, the range of the stroke 
coordinates can have a high variance.
% For instance, the range of coordinates of an electronic board 
% might be very different from those of an smartphone.
In order to deal with such variance, we scale each 
graphic so that the largest dimension of its bounding box
is equal to a fixed parameter $L$ (keeping the original aspect ratio). 
In order to avoid loosing precision, this scaling is done 
coordinate-wise. Once a graphic is scaled, we draw its 
traces through linear interpolations between each 
pair of successive coordinates. The resulting images are gray-scale 
images, with different aspect ratios (but with their largest 
dimension equal to $L$). Figure~\ref{fig:offline} shows some images 
generated through this process.

To generate the bounding boxes, 
we extract the most top-left and bottom-right 
coordinates of each symbol after applying the scaling 
process described above. As the bounding box coordinates 
are measured in terms of pixel units,
after scaling, some boxes might end up having zero 
width or height. In such cases, we update the coordinates 
so that the boxes have a minimum dimension of three pixels,
which is about the width of the drawing traces.

\subsection{Faster R-CNN}
\label{sec:faster}
In this section, we describe the main components of the algorithm 
and highlight parameters of interest regarding our evaluation.
A more detailed description can be found in~\cite{Ren:2015}. 

The algorithm can be seen as a neural network composed 
of three components: feature extractor, Region Proposal Network (RPN), 
and region classifier. The first component receives an input image 
and extracts a feature map, the RPN receives the feature map and 
generates bounding box coordinates (regions) that might 
contain an object, and the region classifier 
classifies the boxes using the features map cropped 
according to the box coordinates.  
The whole network can be trained using 
stochastic gradient descend~\cite{Huang:2017}, or using an iterative process
(iterations of separated RPN and region classifier training
steps)~\cite{Ren:2015}. 

Next sections give more details about the main components.

\subsubsection{Feature extractor}
The feature extractor is a DCNN, usually without 
fully connected layers, that 
maps an input image to a feature map. 
% This CNNs usually consider 
% well known CNNs but considering only until one of the 
% last convolutional layers (avoiding the typical fully 
% connected layers in order to have some localization).
For instance, in~\cite{Ren:2015} the authors 
use a VGG-16~\cite{Simonyan:2015} network and extract 
feature maps from the last (13th) convolutional 
layer. The kind of feature extractors 
determine a large part of the accuracy and computational 
cost of the network.
For instance, in~\cite{Huang:2017}, the authors report 
that inference time varies from about 100 miliseconds when using 
small feature extractors (e.g. Inception V2) to almost 1 second when using 
more complex or deeper DCNNs (e.g. Inception Resnet v2).

The feature extractor allows images with variable width and 
height, but applies a preprocessing step that consists on 
scaling the images so that their minimum dimension ($M$) 
is set to a constant value.

\subsubsection{Region Proposal Network}
The regions proposal network is a two layer 
fully convolutional neural network. The network considers 
a set of boxes, called anchor boxes, of different 
aspect ratios and scales.
For each feature map position and for each anchor box, 
the network outputs the probability of the anchor box 
containing an object. Also, for each anchor box, 
it calculates the coordinates of the box 
that contains the object. The network is 
optimized using a loss function composed of 
a softmax loss for the 
probability outputs, and a regression loss for 
the box coordinates.

An important parameter of the RPN is the number of 
proposals (bounding boxes) that are sent to the classifier. 
The larger the number, the higher the probability of 
finding an object, but also the higher the computational cost 
(and and so the number of false positives) as each region is later 
classified by the region classifier.
The authors in~\cite{Ren:2015} then apply a non-maximum suppression 
algorithm to reduce the number of proposals.
Experiments have shown that $300$ is an adequate number 
when dealing with the PASCAL VOC 2012 dataset.

\subsubsection{Region classifier} The proposals 
generated by the RPN are used to crop the corresponding regions 
from the features map. The cropped regions are then 
used as input to a small neural network classifier 
that determines the class of the object (including 
a background or false positive 
class) and a box refinement. Similar to the RPN, 
this network also uses a softmax and a regression
loss for optimization.

\section{Experimentation}
\label{sec:experimentation}
\subsection{Experimental setup}
We evaluated the methods in the detection of symbols in 
online handwritten mathematical expressions and flowcharts. 
In the first 
case, we used the CROHME-2016~\cite{chrome:2016} dataset, 
and in the second, we used the flowchart dataset proposed 
in~\cite{Awal:2011}. The CROHME-2016 dataset contains 
about $12, 000$ mathematical expressions, and 
the flowcharts dataset contains about 400 flowcharts.
The datasets contain several challenges for the 
detection framework. The CROHME-2016 dataset contains 
a large number of symbol classes (101), including digits, 
characters, operators (e.g. +, $\times$, $\sqrt{\phantom{x}}$). 
Among all classes, specially difficult ones might be the 
small symbols (as \textit{points} and \textit{commas}), 
and symbols that have similar shapes (e.g. 1, |, $\times$, 
x, c and C). The 
flowchart dataset contains seven symbol classes: 
\textit{arrow, text, decision, connection, data, process} 
and \textit{terminator}. 
In such dataset, specially difficult ones are
\textit{texts} and \textit{arrows}, as 
they do not have a specific shape. For instance, \textit{text}  
might consist of a single character, or several words 
placed over several baselines; arrows might be 
horizontal, vertical, or curved lines, and extended 
over a large area of the image. 
Some examples of both datasets are shown in 
Section~\ref{sec:results}.
The datasets are publicly available.

In both mathematical expression and flowchart 
datasets, we used a typical training-validation-test split.
For the test part, we used the same examples defined 
by the dataset authors~\cite{chrome:2016,Awal:2011}. 
For the training and validation parts, 
we randomly selected $80\%$ of the graphics for 
training and used the rest as validation set.

In our implementations, we used the object detection  
framework proposed in~\cite{Huang:2017}. 
As explained in Section~\ref{sec:faster}, a key component 
that determines the algorithm efficiency and accuracy 
is the feature extractor. 
To measure the impact of feature extractors on handwritten 
data, we use four DCNNs of different complexities.
The considered DCNNs, from the smaller to the largest one, 
are: Inception V2~\cite{Ioffe:2015}, 
Resnet 50~\cite{He:2016}, Resnet 101~\cite{He:2016}, 
and Inception Resnet v2~\cite{Szegedy:16}. 
We run the experiments on a Nvidia GeForce GTX Titan 
X GPU 12GB card.

To determine adequate hyper parameters, 
we first experimented with different configurations of the algorithm 
using Inception V2 (as such networks provide a faster 
feedback). In such experimentation we used some default parameters 
from the authors of the algorithm~\cite{Ren:2015}, as well as base code
from~\cite{Huang:2017}.
% First, we use the simplest 
% DCNNs (Inception V2) to set most of the 
% hyper-parameters, as such networks provide a faster 
% training and evaluation. We used the mathematical 
% expression training and validation sets for such evaluation.
From that experimentation, we defined the following 
hyper parameters for all models:   
% In order to define adequate configurations of the 
% Faster R-CNN algorithm, first, we use a small 
% feature extractor 
% Due to the different nature of our images 
% compared to the natural images in which the 
% methods was proposed, we evaluate different 
% configurations to find promising setups, and 
% then applied such configurations to training larger models.
% Using the Inception v2 feature extractor, we 
% evaluate different Faster R-CNN configurations
% and select the best one considering mean average 
% precision (specifically, mAP@0.5) over the 
% CROHME-2016 validation set. The 
% configurations consider:
\begin{itemize}
% \item \textit{Data augmentation.} We consider 
% the use or not of 
\item \textit{Generated images size:} We set the maximum 
image dimension ($L$) to 768.

\item \textit{Scaled images size.} In flowcharts, we set the minimum 
image dimension ($M$) to 600. In mathematical expressions, 
we set $M$ to 300. Although larger values tend to improve 
accuracy~\cite{Huang:2017}, in mathematical expressions, 
we have several cases where images have a very large width, but 
small height. Scaling relative to the height of such images 
generated images with a resolution larger than the one 
allowed by the GPU capacity (when training the models 
with the largest DCNNs). 

\item \textit{Training from pre-trained models.} The base code 
released by~\cite{Huang:2017} includes Faster R-CNN models trained 
over the MSCOCO dataset. 
Although our generated images are very different in 
comparison to the \textit{natural} images of the MSCOCO dataset, 
we found that training using the pre-trained models 
allows for much faster convergence than training 
from scratch. We then used pre-trained models 
for the rest of the experiments.

% \item \textit{Total Stride.} We used stride 16 on all CDNNs 
% with exception of Inception Resnet v2. For the last, we used 
% stride 8 (we tested that stride on smaller DCNNs without 
% improvements on mAP).

\item \textit{Number of proposals.} Once trained, 
we evaluated models that extract from 300 (the default value 
defined in~\cite{Ren:2015}) up to 1000 proposals from the 
RPN. We did not find considerable improvements when using larger 
number of proposals. We then fixed the number of proposals to 300.

\item \textit{Training scheme.} We used minibatch training 
with batch size 1 (due to the variable dimensions of the images). 
We fixed the number of training steps to 
$25, 000$ for flowcharts and $150, 000$ for mathematical 
expressions.

% \item \textit{Feature extractor.} To evaluate speed 
% and accuracy trade-offs, we use feature extractors 
% of different sizes:Inception v2, Resnet 50 and Resnet 101.
\end{itemize}
% In all experiments, we used images 
% with maximum side 768 (over informal experiments, 
% we found that image side to be enough to visually 
% recognize most images), scales [0.25, 0.5, 1.0, 2.0], 
% aspect ratios [0.5, 1.0, 2.0] $150, 000$ steps and mini 
% batch size 1.
% We then use the best configuration to 
% compare the performance of the detection 
% method using larger feature extractors: 
% [REFS].

Additional details about the configuration 
parameters will be available on the code repository.

We used mean average precision (specifically, mAP@0.5) 
as evaluation metric~\cite{Everingham:2010}.

\subsection{Results}
\label{sec:results}
\noindent
\textbf{Flowcharts.}
Table~\ref{tab:flowcharts} shows the detection accuracy of the 
evaluated models over the validation and test sets. 
For the validation set, we show the model's average precision 
using DCNNs with increasing complexities, with the smallest  
one on top of the table.
We can see a consistent improvement as the feature extractors 
are deeper. This improvement is mainly due to higher scores in 
the detection of \textit{texts} and \textit{arrows}. 
For the test set, 
we show the performance of the best model 
(Inception resnet v2) considering mAP. 
We can see that the largest 
variance in accuracy in comparison to the validation set occurs in the detection of \textit{texts} and \textit{arrows}.
% On the test set, we obtained 
% X mAP@0.5. Table Y shows the mAP@0.5 per symbol class.
% Also, we include results of the model using smaller 
% feature extractors.

\begin{table*}[!t]
\renewcommand{\arraystretch}{1.3}
% if using array.sty, it might be a good idea to tweak the value of
% \extrarowheight as needed to properly center the text within the cells
\caption{Average precision on the flowchart validation and test sets}
\label{tab:flowcharts}
\centering
% Some packages, such as MDW tools, offer better commands for making tables
% than the plain LaTeX2e tabular which is used here.
\begin{tabular}{lcccccccc}
\hline
Feature extractor & \multicolumn{8}{c}{Average Precision (\%)}\\
 & mAP & text & arrow & connection & data & decision & process & terminator\\
\hline
\hline
\multicolumn{9}{l}{On validation set:} \\
Inception v2 &  98.6 & 97.3 & 94.7 & 97.9 & 100.0 & 100.0 & 100.0 & 100.0\\
Resnet 50 &  99.2 & 98.3 & 96.0 & 100.0 & 100.0 & 100.0 & 100.0 & 100.0\\
Resnet 101 & 99.5 & 99.3 & 97.6 & 99.7 & 100.0 & 100.0 & 100.0 & 100.0\\ 
Inception Resnet v2 & 99.6 & 99.4 & 97.9 & 100.0 & 100.0 & 100.0 & 100.0 & 100.0 \\
\hline
\hline
\multicolumn{9}{l}{On test set:} \\
Inception Resnet v2 & 97.7	& 95.2	& 91.5 & 99.0 & 99.7	& 99.6 & 99.1 & 99.9\\ 
\hline
\end{tabular}
\end{table*}

\begin{figure*}[!t]
\centering
\subfloat[]{\includegraphics[width=0.45\linewidth]{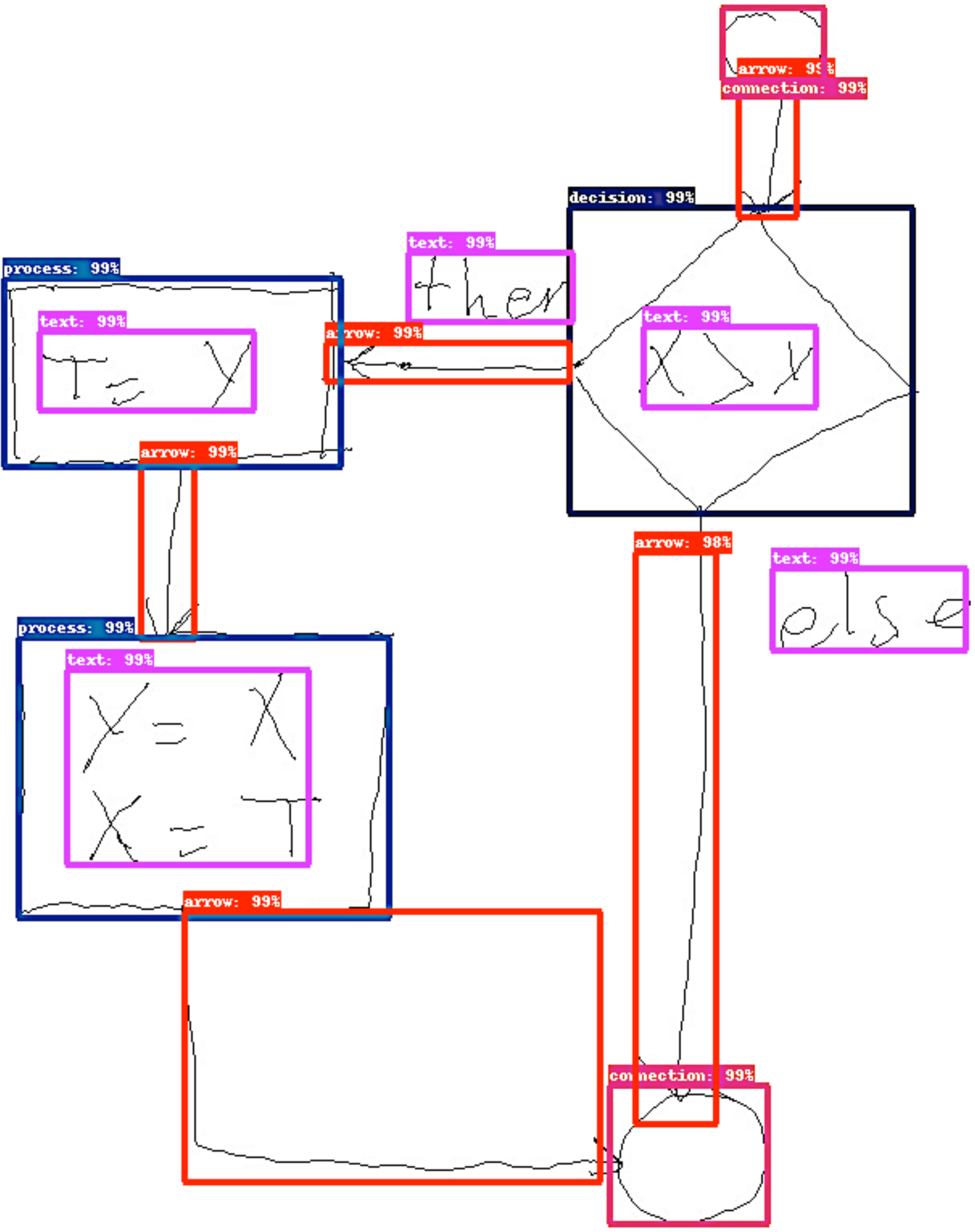}}
\hfil
\subfloat[]{\includegraphics[width=0.4\linewidth]{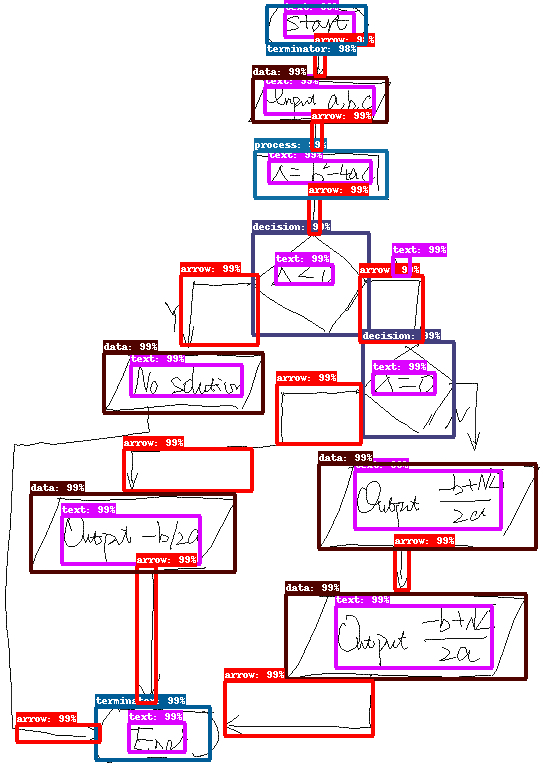}} 
% \subfloat[]{\includegraphics[width=0.8\linewidth]{inv-image-42}}
% \hfil
% \subfloat[]{\includegraphics[width=0.4\linewidth]{inv-image-69}}
\caption{Flowchart output examples. Scores for each detected 
symbol are given by the softmax output (\%).}
\label{fig:flowcharts}
\end{figure*}

Through visual analysis of the output detections,
we found that the most frequent missed symbols are 
\textit{arrows} with several curves and small 
\textit{texts}. Figure~\ref{fig:flowcharts} shows two 
output examples with some miss detection cases.
% Small objects have been described as one of the 
% most challenging objects in object 
% detection methods~\cite{Huang:2017}. 
% In the case of arrows, more data would certainly be helpful 
% to detect arrows with large variance.

\noindent
\textbf{Mathematical expressions.}
% Table~\ref{tab:inception} shows 
% \begin{table}[!t]
% % increase table row spacing, adjust to taste
% \renewcommand{\arraystretch}{1.3}
% % if using array.sty, it might be a good idea to tweak the value of
% % \extrarowheight as needed to properly center the text within the cells
% \caption{Detection results on the CHROHME-2016 validation set. 
% All models use Inception v2 as feature extractor.}
% \label{tab:inception}
% \centering
% % Some packages, such as MDW tools, offer better commands for making tables
% % than the plain LaTeX2e tabular which is used here.
% \begin{tabular}{cccccc}
% \hline
% fine tun. & stride & \#proposals & min. side & mAP@0.5 (\%)\\
% \hline
% \hline
% yes & 16 & 300 & 300 & 83.61\\
% % train. + aug.& yes & 16 & 300 & 300 & Two\\
% no & 16 & 300 & 300 & 78.36\\
% yes & 8 & 300 & 300 & 83.46\\
% yes & 16 & 1000 & 300 & \textbf{83.66}\\
% yes & 16 & 300 & 600 & 82.71\\
% \hline
% \end{tabular}
% \end{table}
Table~\ref{tab:math} shows results on the validation and test 
sets of the CROHME-2016 dataset. We can see higher 
improvements, in comparison to flowchart results, 
as the feature extractors are deeper. Also, by 
analyzing the scores per class, we found that
scores for the most frequent classes are considerably higher
than the mAP score. 
In the same table, we illustrate this by showing the scores for 
the top-10 most frequent classes. These results show that the 
mAP score is pushed down mainly by the less 
frequent classes~\footnote{Recall that mAP is just the mean 
of the average precisions per class~\cite{Everingham:2010}}. 
\begin{table*}[!t]
\renewcommand{\arraystretch}{1.3}
% if using array.sty, it might be a good idea to tweak the value of
% \extrarowheight as needed to properly center the text within the cells
\caption{Average precision (for the top-10 most 
frequent classes and mAP) on the CHROHME-2016 
validation and test sets}
\label{tab:math}
\centering
% Some packages, such as MDW tools, offer better commands for making tables
% than the plain LaTeX2e tabular which is used here.
\begin{tabular}{lccccccccccc}
\hline
Feature extractor & \multicolumn{11}{c}{Average Precision (\%)} \\
& mAP & - & 1 & 2 & + & x & ( & ) & = & a & 3 \\
\hline
\hline
\multicolumn{12}{l}{on validation set:} \\
Inception v2  & 83.6 & 84.8 & 80.6 & 97.6 & 98.7 & 95.3 & 95.4 & 97.3 & 98.4 & 97.8 & 97.7 \\
Resnet 50  & 85.4 & 87.5 & 88.5 & 98.2 & 99.3 & 95.9 & 97.3 & 97.9 & 98.7 & 97.9 & 99.0 \\
Resnet 101  & 87.5 & 92.8 & 91.2 & 98.9 & 98.8 & 96.8 & 97.8 & 98.5 & 98.8 & 97.8 & 99.2\\
Inception resnet v2 & 89.7 & 95.8 & 94.4 & 99.0 & 99.7 & 97.5 & 98.1 & 99.4 & 99.4 & 98.9 & 99.1 \\
\hline
\hline
On test set: & \\
Inception resnet v2 & 86.8 & 96.8 & 92.5 & 99.1 & 99.8 & 98.4 & 99.4 & 99.1 & 99.4 & 95.9 & 99.3 \\
\hline
\end{tabular}
\end{table*}

% Best model map on test set is: $86.77$

% \begin{table}[!t]
% % increase table row spacing, adjust to taste
% \renewcommand{\arraystretch}{1.3}
% % if using array.sty, it might be a good idea to tweak the value of
% % \extrarowheight as needed to properly center the text within the cells
% \caption{Symbol classes with the five highest and 
% lowest mAP@0.5 on CROHME-2016 test set}
% \label{tab:top5}
% \centering
% % Some packages, such as MDW tools, offer better commands for making tables
% % than the plain LaTeX2e tabular which is used here.
% \begin{tabular}{cccc}
% \hline
% \multicolumn{2}{c}{highest mAP}& \multicolumn{2}{c}{lowest mAP} \\
% symbol & mAP@0.5 & symbol & mAP@0.5\\
% \hline
% Inception v2 & 300 & Two & Two\\
% Resnet 50 & 300 & Two & Two\\
% Resnet 101 & 300 & Two & Two\\
% Resnet 50 & 300 & Two & Two\\
% Resnet 101 & 300 & Two & Two\\
% \hline
% \end{tabular}
% \end{table}

\begin{figure}[!htp]
\centering
\subfloat[]{\includegraphics[width=\linewidth]{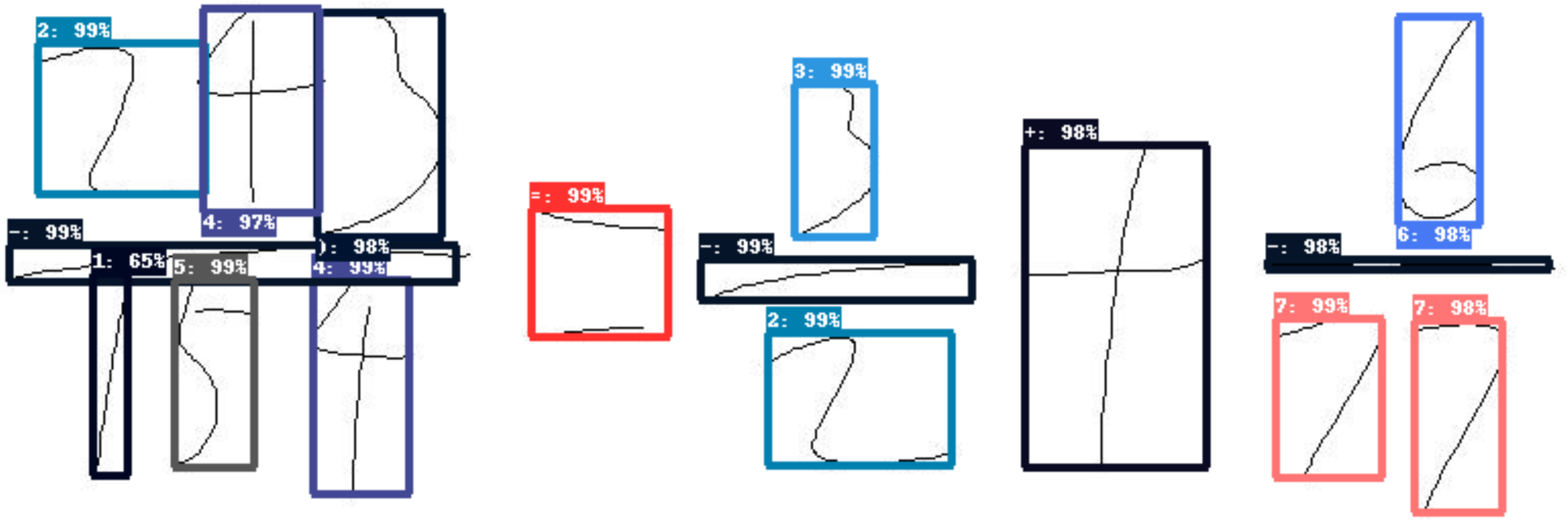}}
\hfil
\subfloat[]{\includegraphics[width=\linewidth]{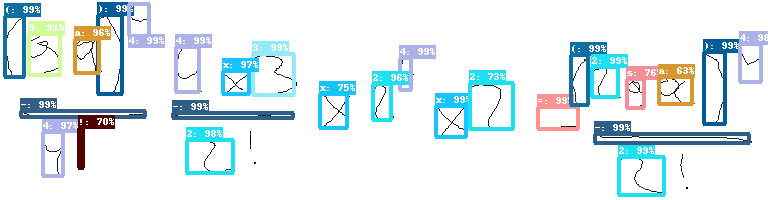}}%
\hfil
\subfloat[]{\includegraphics[width=\linewidth]{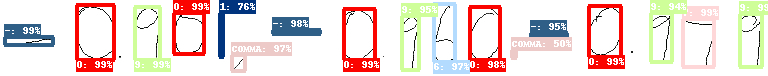}}
\caption{Mathematical expression output examples.}
\label{fig:math}
\end{figure}

In comparison to flowcharts, mathematical expressions 
have several symbol types with a really 
small width or height (e.g. $1$, l, $|$, -, and \textit{dot}).
We found that such symbols are specially difficult 
to be detected by the models. Such difficulty can also be seen 
in the results of the most frequent classes 
in Table~\ref{tab:math}, where the scores of 
\textit{-} and $1$ are low in comparison to the 
scores of the other frequent classes.
Miss classification between symbols that have 
similar shape is other frequent type of error of the detector.
Figure~\ref{fig:math} shows output examples for the 
best model along with some miss detection cases.

\noindent
\textbf{Discussion.} 
It is important to note that although the flowchart 
training data contains only about $200$ examples, 
the data is enough to achieve high accuracy over 
all symbol classes. Furthermore, 
not very deep models, as the Inception v2, already 
allows us to obtain high mAP scores. The possibility of 
using effective and small DCNNs enables 
the use of the method in contexts where computational 
resources are limited or a fast output is required.

Several of the previous works described in 
Section~\ref{sec:related} have reported results on 
our evaluating datasets. For instance, in~\cite{Wang:2016} 
recall of flowchart symbols was $84.4$. 
However, as in such works evaluation is done at stroke level 
and not at bounding box level, results are not 
directly comparable.

\section{Conclusions}
\label{sec:conclusions}
We showed that the Faster R-CNN algorithm 
provides effective detection of symbols in online 
handwritten mathematical expressions and 
flowcharts. Such results are encouraging in the context 
of the development of general methods for symbol 
detection in online handwritten graphics.
Furthermore, the integration of the algorithm 
with structure recognition techniques might 
also accelerate the development of such techniques.

Our evaluation aimed at measuring and understanding the 
potential of the Faster R-CNN algorithm and will 
serve as a baseline for further research. 
We believe that the algorithm  
has high potential for improvement through the introduction 
of online information during the detection pipeline, 
or by solving ambiguities, e.g. using contextual information, 
in a postprocessing or structural recognition step.
\section*{Acknowledgment}
F. D. Julca-Aguilar thanks FAPESP 
(grant 2016/06020-1). N. S. T. Hirata 
thanks CNPq (305055/2015-1). This work is 
supported by FAPESP (grant 2015/17741-9) and 
CNPq (grant 484572/2013-0).



\bibliographystyle{IEEEtran}
\bibliography{bibliography}

\end{document}